\documentclass[lettersize,journal]{IEEEtran}
\usepackage{amsmath,amsfonts}
\usepackage{algorithmic}
\usepackage{algorithm}
\usepackage{array}
\usepackage[caption=false,font=footnotesize,labelfont=rm,textfont=rm]{subfig}
\usepackage{textcomp}
\usepackage{stfloats}
\usepackage{url}
\usepackage{verbatim}
\usepackage{graphicx}
\usepackage{cite}
\usepackage{multirow}
\usepackage{multicol}
\usepackage{color}
\newcommand{\dx}{\textcolor[rgb]{0,0,0}}

\begin{document}

\title{Cross-Modal Adaptive Dual Association for Text-to-Image Person Retrieval}

\author{Dixuan Lin,~\IEEEmembership{}
        Yixing Peng, ~\IEEEmembership{}
       Jingke Meng*, ~\IEEEmembership{}
      Wei-Shi Zheng~\IEEEmembership{}
\thanks{Dixuan Lin, Yixing Peng, Jingke Meng and Wei-Shi Zheng are with the School of Computer Science and Engineering, Sun Yat-sen University, Guangzhou, China.

E-mail:  asxlwsl@gmail.com,  pengyx23@mail2.sysu.edu.cn
, mengjke@gmail.com and wszheng@ieee.org

* Corresponding author.}
}


\maketitle

\begin{abstract}

Text-to-image person re-identification (ReID) aims to retrieve images of a person based on a given textual description. The key challenge is to learn the relations between detailed information from visual and textual modalities. 
Existing works focus on learning a latent space to narrow the modality gap and further build local correspondences between two modalities. 
However, these methods assume that image-to-text and text-to-image associations are modality-agnostic, resulting in suboptimal associations. 
In this work, we show the discrepancy between image-to-text association and text-to-image association and propose CADA: Cross-Modal Adaptive Dual Association that finely builds bidirectional image-text detailed associations. 
  Our approach features a decoder-based adaptive dual association module that enables full interaction between visual and
textual modalities, allowing for bidirectional and adaptive cross-modal correspondence associations. 
Specifically, the paper proposes a bidirectional association mechanism: Association of text Tokens to image Patches (ATP) and Association of image Regions to text Attributes (ARA).
We adaptively model the ATP based on the fact that aggregating cross-modal features based on mistaken associations will lead to feature distortion.
For modeling the ARA, since the attributes are typically the first distinguishing cues of a person, we propose to explore the attribute-level association by predicting the masked text phrase using the related image region.
Finally, we learn the dual associations between texts and images, and the experimental results demonstrate the superiority of our dual formulation.
Codes will be made publicly available.
\end{abstract}


\section{Introduction}
Text-to-image person retrieval is a task that involves retrieving a person of interest from a large image gallery that best matches a given textual description query\cite{li2017person}. Textual descriptions provide a natural and comprehensive way to describe a person's attributes and are more easily accessible than images. As a result, text-to-image person retrieval has received increasing attention in recent years, benefiting a variety of applications from personal photo album searches to public security.

 Comparing to image-based person retrieval methods~\cite{miao2019pose,fan2020correlation,peng2022consistent,ning2020feature,li2021diverse,shi2020person}, text-to-image person retrieval is more challenging due to significant modality heterogeneity caused by inherent representation discrepancies between vision and language. The key to addressing this problem is to establish correspondence between the two modalities. Early methods employed global-matching techniques~\cite{zhang2018deep,zheng2020dual} to pull the image and its corresponding text close together in feature space. However, since global-matching tends to focus on the most salient features, fine-grained information matching is crucial for Re-ID. Recent research has focused on exploring more detailed correspondences between the two modalities, such as image patches and words~\cite{jing2020pose,wang2020vitaa}, and learning local features~\cite{gong2021lag,shao2022learning}.  
{Besides, the knowledge from promising vision-language pre-training models such as CLIP~\cite{radford2021learning} is also exploited and transferred for the text-to-image person retrieval task\cite{jiang2023cross,han2021text,yan2022clip}.}

\begin{figure}[!t]
	\centering
	\centering
 \includegraphics[width=\linewidth]{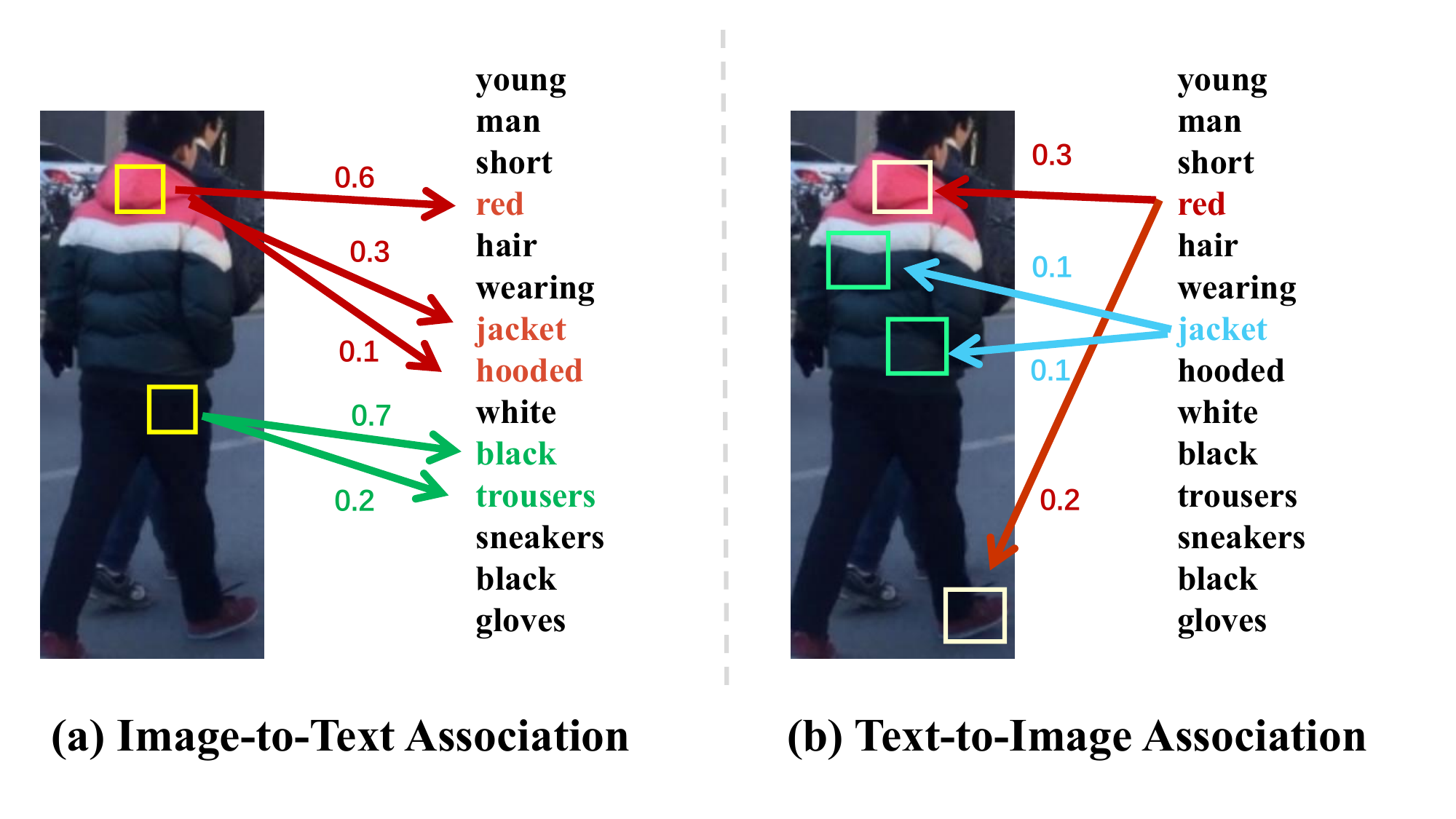}
	\caption{Illustration of {cross-modal dual} association. (a) Associations with image patches as anchors. (b) Associations with textual tokens as anchors.
    Notably, the associations depend on the anchor. For instance, an image patch of a red jacket is highly associated with the word "red" in text, while the word "red" can simultaneously associate with red shoes and a red jacket in the image. }
	\label{fig:multi-c}
\end{figure}

Despite recent advancements in the field, the current methods for building cross-modal associations between {visual} and {textual} are still not entirely satisfactory. The main reason for this is the complexity of establishing detailed correspondences that are relevant to anchors. 
Specifically, if an image patch is taken as an anchor, multiple words could be semantically associated, but the magnitudes of these associations should mainly depend on the image patch. For example, as illustrated in Fig. \ref{fig:multi-c}, an image patch of a red jacket can be associated with the words "red" and "jacket" in the text, while the word "red" can simultaneously be associated with red shoes and red jacket in the image. Additionally, the word "jacket" can associate with all patches the jacket covers. This discrepancy implies that the associations from images to texts do not necessarily imply the associations from texts to images.
Unfortunately, previous works ignored this characteristic and solely adopted one modality as the anchor for learning, resulting in insufficient cross-modal understanding.

In this work, we propose a novel approach called Cross-modal Adaptive Dual Association (CADA). Our approach features a decoder that enables full interaction between visual and textual modalities, allowing for bidirectional and adaptive cross-modal correspondence associations. The dual association implies the  association of the text to image 
 and the  association of the image to text, which is conducive for
the model to understand the corresponding information from two modalities. 
To efficiently learn these dual associations, our decoder shares all parameters with the text encoder except for the cross-attention layers, which are only present in the decoder. 
This is because the self-attention and feed-forward layers work similarly in encoding and decoding tasks, therefore sharing parameters improves training efficiency while enabling dual cross-modal interaction and association learning.
Specifically, the cross-modal adaptive dual association framework utilizes the popular encoder-decoder framework for feature extraction and cross-modal interaction. 
As a foundation for comprehensive and detailed associations, features from the modality-specific encoders are aligned to achieve the {roughly} global-level association (from a sentence to an image).

As the core of our method, we augment a decoder-based adaptive dual association module that enables local-level dual associations. Specifically, the module works on two tasks: the Association of text Tokens to image Patches (ATP) and the Association of image Regions to text Attributes (ARA).
In ATP, we propose to pass information from text tokens to image patches  for building {the token level} associations. 
{We impose a matching constraint to regularize that the built associations should aggregate the image patches related to the text token anchor.}
Moreover, the attribute is usually the first impression when distinguishing a person. Therefore,
we propose ARA to 
explore the association between the text phrase and image patches by proposing Masked Attribute Modeling (MAM). MAM generates a masked text phrase by adaptively locating related image regions, inspired by the success of Masked Language Modeling in vision-language pretraining~\cite{devlin2018bert,li2021align,shu2023see,jiang2023cross}.

In summary, the contributions of this work are three-fold.
\begin{itemize}
	\item [1)]
 \dx{We propose a Cross-modal Adaptive
Dual Association (CADA) approach which bidirectionally associates the visual and textual modalities while current methods treat the two non-equivalent associations as one association.}
	\item [2)]
 {We introduce the Association of text tokens to image patches (ATP), which allows for passing information from the textual modality to the visual modality and adaptively aggregating image patches related to text token anchors.}
	\item [3)]
 {We propose an Association of image regions to text attributes (ARA) to explore the relationship between the text phrase and the image patches, which is achieved by the Masked Attribute Modeling(MAM) to generate a masked text phrase by adaptively locating the related image regions.}
\end{itemize}
We conduct extensive evaluations to verify the effectiveness of our proposed approach 
on three public datasets, i.e., CUHK\cite{li2017person}, ICFG-PEDES\cite{ding2021semantically} and RSTPReid\cite{zhu2021dssl}. 
The results show that the proposed CADA method outperforms state-of-the-art methods by a large margin, demonstrating the effectiveness of our proposed approach.

\begin{figure*}[t]
	\centering
	\includegraphics[width=1.0\textwidth]{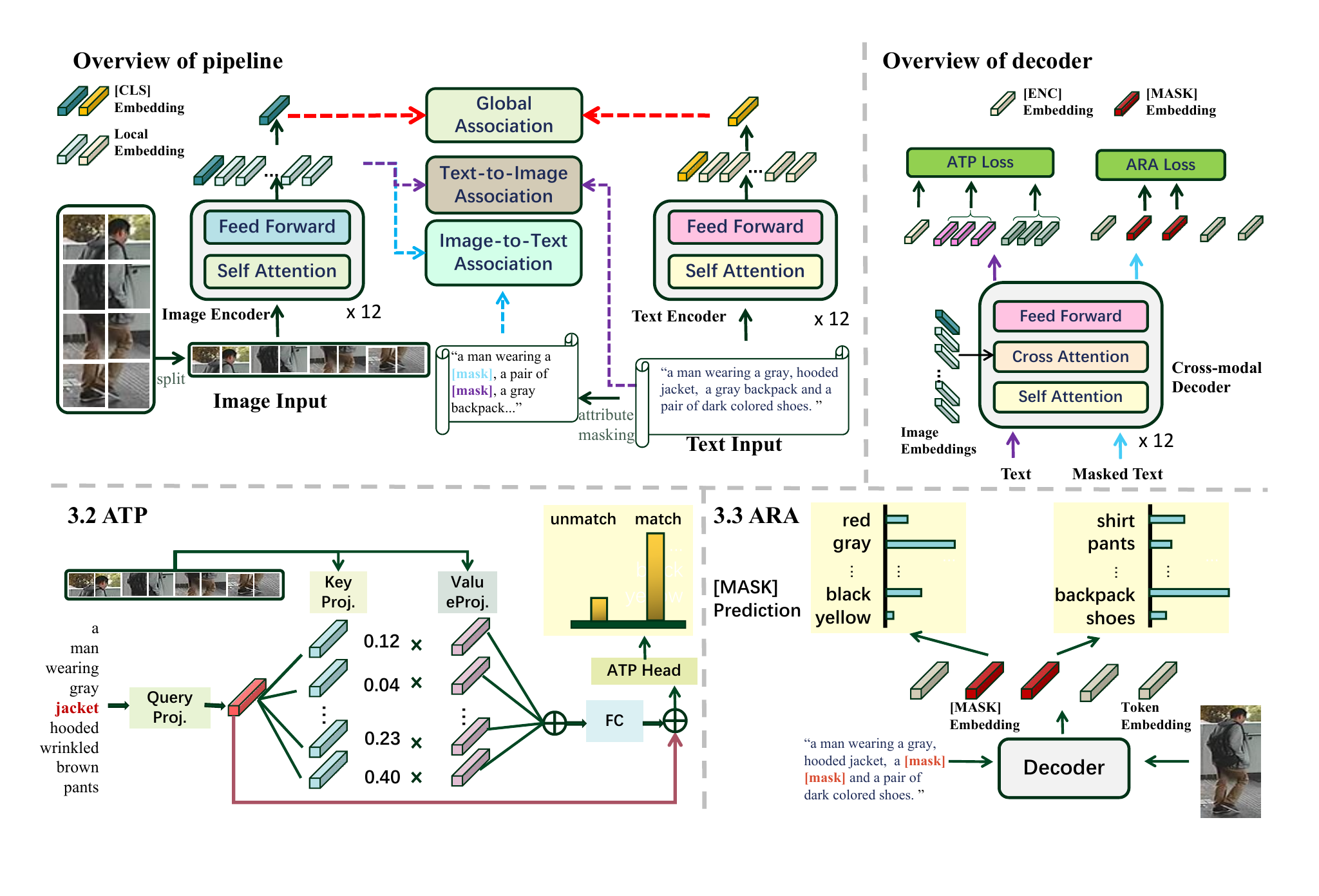}
	\caption{Overview of our CADA framework, which includes two modality-specific encoders and a cross-modal decoder (The same color denotes the shared parameters). Encoders extract global features and learn global association under Normalized Distribution Fitting loss in Sec.\ref{sec:global}. The decoder learns cross-modal dual associations by the Association of Text Tokens to Image Patches (ATP) module and the Association of Text Tokens to Image Patches(ARA) module. In the ATP module, we exploit the cross-attention mechanism to aggregate each token with relevant image patches by learning the text-image matching probability, which essentially guides the model to capture the associations of text tokens to image patches.
    In the ARA module, the model is enforced to restore masked attribute phrases via cues from image patches, which requires the associations of image patches to text tokens.
    Details of ATP and ARA are formulated in Sec.\ref{TTA} and Sec.\ref{IAA} respectively.} 
	\label{modelpic}
\end{figure*}

\section{Related work}
 \subsection{Text-to-Image Person Retrieval} 
 Text-to-image person retrieval was first proposed by \cite{li2017person}, releasing the first benchmark dataset CUHK-PEDES. Later, more challenging datasets ICFG-PEDES \cite{ding2021semantically} and RSTPReid \cite{zhu2021dssl} were {proposed} for text-to-image person retrieval. Compared with general text-image retrieval benchmark datasets e.g. COCO \cite{lin2014microsoft}, and methods \cite{zheng2020dual,li2021align} person retrieval is more challenging for its high demand on cross-modal detail alignment. According to the strategy in alignment, current works can be classified into two branches: cross-modal interaction-free methods and cross-modal interaction-based methods. The former branch of methods \cite{zheng2020dual,wang2019language,liu2019deep} focuses on developing a better feature extraction model to learn discriminative image-text embeddings in a shared latent space, {the CMPM loss  \cite{zhang2018deep} is proposed to}
 align cross-modal representations, {IVT\cite{shu2023see} introduces an additional bidirectional mask modeling module to urge global features to contain multi-level information. } Despite the efficiency of these methods, their performance is not satisfactory because they only align global representations, wasting the abundant detail information in local features. 

Due to the deficiency of global alignment for including overall details, cross-modal interaction-based methods are purposed to fully utilize the local information. 
Recent works focus on local-level correspondence. Some works concentrate on salient patches, \cite{chen2018improving,li2017person,niu2020improving,ji2022asymmetric} purpose patch-word matching framework, Ding~\cite{ding2021semantically} utilizes attention mechanism to capture the relation between body parts. {More recent works drop external cues like segmentation and body landmarks, employing implicit alignment instead. AXM~\cite{farooq2022axm} uses multi-scale convolution layers in feature extraction, TIPCB~\cite{chen2022tipcb} uses different residual blocks to capture semantic information from different scales, then align cross-modal features in the same scale. CAIBC~\cite{wang2022caibc} aligns color-related features and color-irrelated features respectively. CFine ~\cite{yan2022clip} adopts Vision-Transformer~\cite{dosovitskiy2020image} and BERT~\cite{devlin2018bert} as backbones for feature extraction, achieving better performance.}  

{
{
However, the current methods are still not satisfactory in building {local-level} cross-modal associations. 
The detailed correspondences are complicated and relevant to anchors.
The associations from images to texts do not mean the associations from texts to images.
Unfortunately, previous works ignore the characteristic and solely adopt one modality as the anchor for learning, leading to insufficient cross-modal understanding.}

Differently, {we introduce a decoder-based adaptive dual association module that enables comprehensive interaction between visual and textual modalities. This allows for bidirectional and adaptive cross-modal correspondence associations, which are crucial for learning the relation between the two modalities in a detailed and accurate way.}

}
  

\subsection{Vision-Language Pre-training(VLP)}
The process of vision-language pre-training(VLP) is initializing the vision-language model with parameters pre-trained on a large-scale image-text dataset, then finetuning the model on downstream 
 tasks. Many previous works~\cite{he2016deep,dosovitskiy2020image,devlin2018bert} have proven that transferring knowledge in this way can improve the performance of downstream tasks. Some works align global representations of images and texts in a shared space. CLIP \cite{radford2021learning} and ALIGN \cite{jia2021scaling} employ contrastive learning loss to learn strong coherent cross-modal representations for cross-modal alignment. Some other works exploited the interaction of local features. ALBEF \cite{li2021align} combines contrastive learning and masked language modeling (MLM) with a dual-encoder design to learn image-to-text relations. Coca \cite{yu2022coca} introduces a decoder to aggregate cross-modal representations, BLIP \cite{li2022blip} simultaneously utilizes multi-stage learning to improve the ability of image-text understanding and image-based captioning. 

{Recent developments in text-image person retrieval have achieved notable improvements by utilizing pre-trained vision-language models, instead of relying on backbones that were only pre-trained on unimodal datasets.} Recent studies~\cite{yan2022clip,jiang2023cross} propose CLIP-driven frameworks, using a pre-trained CLIP as the backbone and finetuning it on text-to-image person retrieval datasets. 
{Similarly, this work also draws inspiration from the success of VLP models and leverages their knowledge to enhance both global alignment and local interaction in text-to-image person retrieval task.}

\section{Cross-Modal Adaptive Dual Association}

\subsection{{The Architecture Overview}}
\label{sec:global}
In this section, we {introduce the overview architecture of} the proposed Cross-modal Adaptive Dual Association (CADA) model. 
The CADA framework is conducted based on a popular encoder-decoder {network architecture}.
The encoder is for extracting high-level semantic information from data of different modalities, and the decoder enables detailed bidirectional cross-modality interactions, which is the core for effective cross-modality Re-ID.
{As for the encoder, we propose an encoder-based global association module to roughly align the global feature of two modalities. As for the decoder, we propose a decoder-based local adaptive dual association module to enhance the interaction of the visual and textual
modalities in a bidirectional way.
Our method succeeds in adaptively associating the local region of the image and text in the token-level and attribute-level.}
The details are as follows.
\subsubsection{Encoder-based Global Association}
\label{SecNDF}
\ 
\newline
{\bfseries Image encoder.} We utilize a pre-trained ViT as the image encoder to extract visual features.
For a given image $I\in R^{H\times W\times C}$, where $H, W, C$ respectively denote its height, width, and channels, 
we split it into a sequence of $N=H\times W/P^2$ patches, where $P$ represents the length of one patch. 
An additional learnable token $[CLS]_v$ is attached to the beginning of the patch sequence as a global visual representation. Learnable position embeddings are added to incorporate spatial information. 
Then the sequence of patches is fed into the visual transformer. 
The output of the image encoder is represented as $\{v_{cls},v_1,...,v_N\}$, where $v_{cls}$ is a $d_v$-dim embedding of the $[CLS]_v$ token.

\noindent \textbf{Text encoder.}
{ \bfseries } We use the BERT of 12 layers as our text encoder. 
For a given text $T$, we correspondingly map each word in $T$ to its word embedding. 
Special token $[CLS]_t$ is used to represent the whole sentence and $[PAD]$ is used to pad the embeddings to a fixed length. 
The sequence of word embedding aggregated with position embedding is fed into the text encoder to obtain textual representations $\{t_{cls},t_1,...,t_M\}${, where $M$ is the length of sentence and $t_{cls}$ is a $d_t$-dim embedding of $[CLS]_t$ token.}

\noindent{\bfseries Global association.} The global association is learned by aligning global features of image and text in a shared latent space. Previous works~\cite{yan2022clip,chen2022tipcb,wang2022caibc} demonstrate that image-text contrastive learning can effectively align the global feature of two modalities {by the cross-modal projection matching (CMPM) loss~\cite{liu2019deep}}.
{
We notice that CMPM loss takes projection length as similarity, where the magnitude of the vector can be a great impact.
To alleviate the drawbacks of CMPM, we proposed a novel Normalized Distribution Fitting (NDF) loss that applies cosine similarity instead of projection, and uses Forward KL Divergence and Backward KL Divergence to accelerate distribution fitting.} 

Given an image $I$ and a text $T$, we obtain global representations $\tilde{v} = W_vv_{cls}$ and $ \tilde{t} = W_tt_{cls}$ with learnable linear transformations $W_v\in \mathbb{R}^{d\times d_v}$ and $W_t \in \mathbb{R}^{d\times d_t}$.
We define $sim(I,T)=\tilde{v}^\top \tilde{t} / \lvert \tilde{v}\rvert\lvert \tilde{t}\rvert$ as similarity function.
In a mini-batch consisting of $N$ image-text pairs, the image-to-text matching probability of $I$ to $T$ is computed as:

\begin{equation}
    \label{eq:globsim}
    p^{i2t}=\frac{\exp(sim(I,T)/\tau)}{\sum_{k=1}^{N}\exp(sim(I,T_k)/\tau)},
\end{equation}
where $\tau$ is a temperature parameter used to control the sharpness of the probability distribution. 

Assume there are $N_z$ images and texts in a mini-batch. Denoting the ground-truth one-hot label as $q^{i2t} \in \mathbb{R}^N_z$, the image-to-text contrastive loss is formulated:
\begin{equation}
\label{conloss}
    \begin{aligned}
        L_{i2t}(I)&=KL(p^{i2t}||q^{i2t}) + KL(q^{i2t}||p^{i2t}), 
    \end{aligned}
\end{equation}
where $KL(p||q)$ represents KL divergence from distribution $p$ to distribution $q$, focusing on samples where $p$ has high probability while $q$ has low probability. We term the former part in Eq.(\ref{conloss}) as Forward KL divergence and the latter part as Backward KL divergence. When minimizing $L_{i2t}$, Forward KL divergence focuses on pulling down the matching probability of negative pairs, meanwhile Backward KL divergence focuses on pulling up the matching probability of positive pairs. In this way, the alignment process is accelerated.

Similarly, the text-to-image contrastive loss $L_{t2i}(T)$can be formulated by exchanging $I$ and $T$ in Eq.(\ref{eq:globsim}),(\ref{conloss}). Our proposed Normalized Distribution Fitting(NDF) loss can be formulated as:
\begin{equation}
\label{contrastive}
    L_{{NDF}} = \frac{1}{N_z}(\sum_{n=1}^{N_z} L_{i2t}(I_n)+\sum_{n=1}^{N_z} L_{t2i}(T_n)).
\end{equation}

{
\subsubsection{Decoder-based  Local Adaptive Dual Association}
\ 
\newline
The encoder-based global association works on the representations of holistic images and texts, which is insufficient to associate comprehensive and detailed information in two modalities. 
Therefore, we propose  a decoder-based  local adaptive dual association 
scheme to enhance the interaction of the visual and textual modalities 
in a bidirectional way and adaptively associate the local region of the image and text in the token-level and attribute-level.
Specifically, the decoder-based  local adaptive dual association is achieved by the association of text tokens to image patches (ATP) and the association of image regions to text
attributes(ARA). 
We detail the ATP and ARA in Section \ref{TTA} and Section \ref{IAA}.
}

\subsection{{Association of   Text Tokens to Image Patches}}\label{TTA}

For the association of text tokens to image patches (ATP), 
we {pass information from textual modality to visual modality by adaptively aggregating relevant image patches based on text token anchors.}
\dx{Since a single word can relate to different parts in an image, we need to determine which certain part the word is describing based on its position in the sentence. So we conduct self-attention mechanism and cross-attention mechanism to obtain text-image aggregated features.}
Intuitively, if an image and a text are truly matched, the associations between them should be strong therefore the corresponding features can be aggregated naturally.
By contrast, aggregating the features from a negative pair (the image and text refer to different persons) will cause feature distortions.
In light of this, we interact the association between the words and image patches by feedforwarding the text embeddings and image embeddings into the decoder with cross-attention layers.
{By evaluating matching loss on the aggregated features}, the cross-attention layer learns how to match and aggregate token-level fine-grained features from different modalities.


The decoder architecture and cross-attention mechanism in ATP are illustrated in Fig.~\ref{modelpic}. 
Given an image-text pair $(I_a,T_b)$, we first tokenize and embed $T_b$ to get textual embeddings. Special token $[ENC]$ is attached at the beginning of the sequence.
Then token embeddings $E^b=\{e_{enc},e_1,...,e_M\}$ are fed into cross-attention layer. The image embeddings {$V^a=\{v_{cls},v_1,...,v_N\}$} from the image encoder are also feed-forwarded into the cross-attention layer. 
We model the association in the cross-attention layers. When operating cross-attention, $E^b$ is taken as query, and $V^a$ is taken as key and value. 
We denote $H^{a,b}=\{h^{a,b}_{enc},h^{a,b}_{1},...,,h^{a,b}_{M}\}$ as the final output {embeddings} of decoder. 
\dx{The output embedding $h^{a,b}_{i}$ is obtained by aggregating $e_i$ with weighted combination of image embeddings in $V^a$, where relevant image patch gets a higher weight. If a token can correspond with certain image patches, its aggregated embedding shall be harmonious, otherwise confusing.}
To guarantee the related  image features can be aggregated naturally corresponding to the text token, the most straightforward  way is to constrain each element in $H^{a,b}$ with a binary classification loss to predict whether a text token and its corresponding aggregated feature is positive
(matched) or negative (unmatched). 
Considering the training efficiency and noises from meaningless words (e.g., "a", "of", and so on), we split $H^{a,b}$ into groups and calculate matching loss for each group.
Specifically, we split the aggregated features $H^{a,b}$ with group size $p$ and slitting stride $r$ into $\kappa$ groups:$\{G^{a,b}_1,...,G^{a,b}_\kappa\}$, where $\kappa = \frac{(M-p)}{r}+1$. 
The $i$-th group is denoted as $G^{a,b}_i = \{h^{a,b}_{(i-1)\times r},...,h^{a,b}_{(i-1)\times r + p}\}$.
We apply mean pooling to each group $G^{a,b}_i$ and obtain $g^{a,b}_i\in \mathbb{R}^d$ as a group representation.
In addition, we take $g^{a,b}_0 = h^{a,b}_{enc}$ as an extra group. 
For each group representation $g^{a,b}_i (0<i<\kappa$), we feed it into a classifier to compute the probability of "match" and "mismatch", formulated as: 
\begin{equation}
    \label{eq:localsim}
    \hat{p}(g^{a,b}_{i})=Softmax(FC_{\phi}(g^{a,b}_{i})),
\end{equation}
where $\hat{p}(g^{a,b}_i)$ is the matching probability of $I_a$ and $T_b$ based on $g^{a,b}_i$. 
We expect to get a high matching probability for positive pairs and get a low matching probability for negative pairs. 
For each $I_a$ we select one hard negative text $T_{a^-}$ in the mini-batch which has highest global similarity $sim(I_a,T_{a^-})$ while $I_a, T_{a^-}$ refer to different persons. For each $T_b$, we select $I_{b^-}$ in the same way. 
Overall, the ATP loss is computed as:
\begin{equation}
\begin{aligned}
    L_{ATP} = \frac{1}{\lvert \mathcal{P} \rvert * (\kappa+1)} \sum_{(I_a,T_b)^+\in \mathcal{P}} \sum_{i=0}^{\kappa} [ &  \log(\hat{p}(g^{a,b}_i))\\ + \log(1-\hat{p}(g^{a,a^-}_i))  & + \log(1-\hat{p}(g^{b^-,b}_{i}))].
\end{aligned}
\end{equation}
where $\mathcal{P}$ is the set of positive pairs in a mini-batch. $(I_a,T_b)^+$ denotes a positive image-text pair where $I_a$ and $T_b$ refer to a same person. By minimizing $L_{ATP}$, the model learns to distinguish negative image-text pairs only discrepant in details, forcing it to correctly build the association between text tokens and image patches.

\subsection{Association of  Image Regions to Text Attributes}\label{IAA}

\dx{In ATP module our model learns to understand association from words in text to image patches. In this section we discuss the image-to-text association.}
To learn the image-to-text association, we propose an 
ARA module that focuses on the association of image regions to text attributes. Since the attribute is typically the first thing that comes to mind when identifying a person, it is an important feature to explore for attribute-level cross-modal association. To achieve this, we propose the Masked Attribute Modeling (MAM) technique, inspired by the success of masked language modeling in vision-language pre-training~\cite{devlin2018bert,yang2022vision,li2021align,jiang2023cross}. In MAM, we randomly mask attribute phrases in a textual description, and the decoder is trained to predict the correct  phrase by locating the related image region. 

Compared with MLM, MAM focuses on attribute-related words rather than randomly masking out all words, 
{since the attributes in the image are important for person image retrieval. }
To achieve this, we use a natural language toolkit NLTK\footnote{https://www.nltk.org/} to conduct part-of-speech tagging, then denote phrases in the pattern of "[adj][noun]" as attributes, such as "black shoes" and "long straight hair".

Masking out an attribute phrase means each word in that phrase is replaced by $[MASK]$. The decoder learns to predict the whole masked attribute phrase. Specifically, given a positive image-text pair $(I_a,T_b)$, we obtain masked text $T_{b'}$ by randomly masking attribute phrases with a masking rate $\alpha$. We then feed $(I_a,T_{b'})$ into decoder to obatin final embeddings $H^{a,b'}=\{h^{a,b'}_{enc},h^{a,b'}_{1},...,,h^{a,b'}_{M}\}$ as in Sec.\ref{TTA}. We denote $H^{a,b'}_{msk}\in \mathbb{R}^{N_m\times d}$ as a subset of $H^{a,b'}$, consisting of final embeddings of every [MASK] token in $T_{b'}$. $N_m$ denotes the number of masked tokens in $T_{b'}$.
The prediction process can be formulated as a classification problem. Following BERT~\cite{devlin2018bert}, we hold a vocabulary of size $Voc$, including all possible words.
We feed $H^{a,b'}_{msk}$ into a classifier to compute the probability for each word in vocabulary:
\begin{equation}
   \label{maskPred} p_{msk}^{a,b'}=Softmax(FC_{\beta}(H^{a,b'}_{msk})),
\end{equation}
where $ p^{a,b'}_{msk} \in \mathbb{R}^{N_m\times Voc}$ is the predictions for every masked token. 
We take the label of masked words $y_{mask}^{a,b'}$ from the vocabulary as the ground-truth label, and the ARA loss  is computed as:
{
\begin{equation}
    L_{ARA} =\frac{1}{\lvert \mathcal{P} \rvert}\sum_{(I_a,T_b)^+\in \mathcal{P}} KL(y_{msk}^{a,b'}||p_{msk}^{a,b'}),
\end{equation}
where $\mathcal{P}$ is the set of positive image-text pairs in a mini-batch. We apply KL divergence to align predicted distribution and ground-truth distribution.} 
The ARA module achieves image-to-text associations by predicting correct attributes using the related image regions.

\subsection{Training and Inference}
\noindent {\bfseries Training.}
In summary, the overall loss of {our Cross-modal Adaptive
Dual Association (CADA)  framework} can be formulated as:
\begin{equation}
\label{eq:loss}
    L_{CADA} = \lambda L_{NDF} + L_{ATP} + L_{ARA},
\end{equation}
where $\lambda$ is a trade-off parameter. 

\noindent{\bfseries Inference.} 
\label{inference_method}
{During the testing stage, we conduct the evaluation under the global-matching and local-matching inference protocols which have been defined in previous work~\cite{jiang2023cross}.\\
\textbf{The global-matching inference}: we separately extract global features by image encoder and text encoder, then utilize the cosine similarity $S_G$ between global features (Eq.(\ref{eq:globsim})) as the final matching score.\\
\textbf{The local-matching inference}: image and text feature sequences are fed into the decoder to interact and  compute matching score $S_L$ (i.e., $\hat{p}(g^{a,b}_{0})$ in Eq.(\ref{eq:localsim})).
For efficiency, we  rank all images in the gallery based on $S_G$ and  select the top $\eta$ candidates to compute the $S_L$. We take $S_G + S_L$ as the final matching score of the top $\eta$ candidates and $S_G$ as the final matching score of the rest. 
We set $\eta=32$ when testing on all benchmarks. The performance under different $\eta$ is discussed in Sec.\ref{k_test}. 
}

\section{Experiments}

\subsection{Datasets and Evaluation Metrics}
\noindent{\bfseries Datasets.}
We evaluate our methods on three public datasets including  \textbf{CUHK-PEDES}\cite{li2017person}, \textbf{ICFG-PEDES}\cite{ding2021semantically}, and \textbf{RSTPReid}\cite{zhu2021dssl}.

CUHK-PEDES is the first accessible dataset for text-image person retrieval. It includes 40,206 images and 80,412 text descriptions of 13,003 identities, each image is annotated with 2 textual descriptions. Following the official protocol, the dataset is split
into 34,054 images and 68,108 descriptions for 11,003 identities in the training
set, 3,078 images with 6,156 descriptions for 1,000 identities in the validation set, and 3,074 images with 6,148 descriptions of 1,000 persons in the testing set.

ICFG-PEDES is collected from MSMT-17 dataset\cite{wei2018person}, it has more identities and textual descriptions. ICFG-PEDES contains 54,522 text descriptions for 54,522
images of 4,102 persons. It is split into 34,674 image-text pairs of 3,102 identities in the training set and 19,848 image-text pairs of 1,000 identities in the testing set.

RSTPReid is identically collected from MSMT-17 including 20,505 images with 41,010 textual descriptions of 4,101 persons. Each person contains 5 images captured by 15 different cameras, and each image has 2 corresponding textual descriptions. The training contains 3,701 identities, and the validation set and testing set contains 200 and 200 respectively. 

\noindent{\bfseries Evaluation Metrics.}
For all experiments, we report Rank-1, Rank-5, and Rank-10 accuracy and mean Average Precision (mAP) to evaluate the performance of our model.

\subsection{Implementation Details}
We use ViT-B/16 \cite{dosovitskiy2020image} of 12 layers as the image encoder and BERT-base \cite{devlin2018bert} of 12 layers as the text encoder. 
We initialize our model with parameters of BLIP~\cite{li2022blip}, pre-trained on 129M image-text pairs. 
During training, we employ random horizontal flipping, random erasing, and random crop for data augmentation. Each image is resized to $224\times 224$. The input sentences are set with a maximum text length of 72. The embedding dimension $d_v, d_t =768$, and the shared latent dimension $d=256$. The masking rate $\alpha$ of MAM is 0.8,
each masked word is replaced by a special token $[MASK]$. The temperature parameter $\tau$ for the NDF loss in Eq.(\ref{eq:globsim}) is set to 0.02. 
Parameter $\lambda_{NDF}$ in Eq.(\ref{eq:loss}) is set to 0.1. We use the AdamW optimizer~\cite{kingma2014adam} with a weight decay of 0.05. Our model is trained with a batch size of 96 and lasted for 40 epochs. The learning rate is initialized to $1e-5$, with the cosine learning rate decay scheduler.

\begin{table}[htbp]
\caption{Performance comparisons on the CUHK-PEDES dataset. "G" and "L" in "Type" denote global-matching and local-matching {inference protocols}. 
}
\label{cuhksota}
\begin{center}
\resizebox{\linewidth}{!}{
\begin{tabular}{c|cc|cccc}
\hline
Methods  &Type & Ref      & Rank-1 & Rank-5 & Rank-10 & mAP                  \\
\hline\hline
Dual Path\cite{zheng2020dual}&G & TOMM20   & 44.40  & 66.26  & 75.07   & -                    \\
CMPM/CMPC\cite{zhang2018deep}&L & ECCV18   & 49.37  & -      & 79.27   & -                    \\

MIA \cite{niu2020improving}  &L    & TIP20    & 53.10  & 75.00  & 82.90   & -                    \\
A-GANet\cite{liu2019deep}  &G & MM19     & 53.14  & 74.03  & 81.95   & -                    \\

SCAN \cite{lee2018stacked} &L    & ECCV18   & 55.86  & 75.97  & 83.69   & -                    \\
ViTAA \cite{wang2020vitaa}  &L  & ECCV20   & 55.97  & 75.84  & 83.52   & 51.60                \\

NAFS \cite{gao2021contextual} &L    & arXiv21  & 59.94  & 79.86  & 86.70   & 54.07                \\
DSSL \cite{zhu2021dssl}  &L   & MM21     & 59.98  & 80.41  & 87.56   & -                    \\

SSAN  \cite{ding2021semantically} &L   & arXiv21  & 61.37  & 80.15  & 86.73   & -                    \\
LapsCore \cite{wu2021lapscore} &L & ICCV21   & 63.40  & -      & 87.80   & -                    \\
IVT \cite{shu2023see}   &G   & ECCVW22  & 64.00  & 82.72  & 88.95   & -                    \\
LBUL \cite{wang2022look}  &L   & MM22     & 64.04  & 82.66  & 87.22   & -                    \\
TextReID \cite{han2021text} &G & BMVC21   & 64.08  & 81.73  & 88.19   & 60.08                \\
SAF \cite{li2022learning}  &L  & ICASSP22 & 64.13  & 82.62  & 88.40   & -                    \\
TIPCB \cite{chen2022tipcb} &L   & Neuro22  & 64.26  & 83.19  & 89.10   & -                    \\
CAIBC \cite{wang2022caibc}  &L    & MM22     & 64.43  & 82.87  & 88.37   & -                    \\
AXM-Net \cite{farooq2022axm} &L & AAA122   & 64.44  & 80.52  & 86.77   & 58.73                \\
LGUR \cite{shao2022learning} &L &MM22 &65.25 &83.12 &89.00& - \\
ACSA \cite{ji2022asymmetric} &L & TMM22 &68.67 &85.61 &90.66 & - \\
CFine  \cite{yan2022clip} &G  & arXiv22  & 69.57  & 85.93  & 91.15   & -                    \\
IRRA \cite{jiang2023cross}  &G   & CVPR23   & 73.38  & 89.93  & 93.71   & 66.13                \\
\hline
\textbf{CADA-G(Ours)} &  G   & -    &73.48          & 89.57          & 94.10          & 65.82  \\                   
\textbf{CADA-L(Ours)} &  L   & - &   \textbf{78.37} & \textbf{91.57} & \textbf{94.58} & \textbf{68.87}\\
\hline
\end{tabular}}
\end{center}
\end{table}
\subsection{Comparison with Other Methods}
We compare our method with state-of-the-art methods on three benchmark datasets as shown in Table \ref{cuhksota}, Table \ref{icfgsota}, and Table \ref{rstpsota}. 

\begin{table}[htbp]
\begin{center}

\caption{Performance comparisons on ICFG-PEDES. 
}
\label{icfgsota}
\resizebox{\linewidth}{!}{
\begin{tabular}{c|cc|cccc}
\hline
Methods  &Type & Ref     & Rank-1 & Rank-5 & Rank-10 & mAP   \\
\hline\hline
Dual Path \cite{zheng2020dual}&G & TOMM20  & 38.99  & 59.44  & 68.41   & -     \\
CMPM/CMPC \cite{zhang2018deep}&L & ECCV18  & 43.51  & 65.44  & 74.26   & -     \\
MIA    \cite{niu2020improving} &L    & TIP20   & 46.49  & 67.14  & 75.18   & -     \\
SCAN  \cite{lee2018stacked} &L   & ECCV18  & 50.05  & 69.65  & 77.21   & -     \\
ViTAA  \cite{wang2020vitaa} &L  & ECCV20  & 50.98  & 68.79  & 75.78   & -     \\
SSAN   \cite{ding2021semantically}&L   & arXiv21 & 54.23  & 72.63  & 79.53   & -     \\
TIPCB  \cite{chen2022tipcb} &L   & Neuro22 & 54.96  & 74.72  & 81.89   & -     \\
IVT    \cite{shu2023see}  &G   & ECCVW22 & 56.04  & 73.60  & 80.22   & -     \\
CFine  \cite{yan2022clip} &G  & arXiv22 & 60.83  & 76.55  & 82.42   & -     \\
IRRA   \cite{jiang2023cross} &G   & CVPR23  & 63.46  & 80.25  & 85.82   & 38.06\\
\hline
\textbf{CADA-G(Ours)} &  G   &  -   & 62.54 & 79.46 & 85.14 & 37.07                    \\
\textbf{CADA-L(Ours)} &  L   &  -   &  \textbf{67.81} & \textbf{82.34} & \textbf{87.14} & \textbf{39.85}\\
\hline
\end{tabular}}
\end{center}
\end{table}

\begin{table}[!htbp]
\begin{center}

\caption{Performance comparisons on RSTPReid. 
}
\label{rstpsota}
\resizebox{\linewidth}{!}{
\begin{tabular}{c|cc|cccc}
\hline
Methods  &Type & Ref     & Rank-1 & Rank-5 & Rank-10 & mAP   \\
\hline\hline
DSSL \cite{zhu2021dssl}  &L   & MM21    & 39.05  & 62.60  & 73.95   & -     \\
SSAN  \cite{ding2021semantically} &L & arXiv21 & 43.50  & 67.80  & 77.15   & -     \\
LBUL \cite{wang2022look} &L   & MM22    & 45.55  & 68.20  & 77.85   & -     \\
IVT   \cite{shu2023see} &G & ECCVW22 & 46.70  & 70.00  & 78.80   & -     \\
ACSA \cite{ji2022asymmetric} &L & TMM22 & 48.40 &71.85 &81.45 & - \\
CFine \cite{yan2022clip}&G  & arXiv22 & 50.55  & 72.50  & 81.60   & -     \\
IRRA  \cite{jiang2023cross}&G   & CVPR23  & 60.20  & 81.30  & 88.20   & 47.17\\
\hline
\textbf{CADA-G(Ours)} &   G &   -   &    61.50 & 82.60 & 89.15 & 47.28                    \\
\textbf{CADA-L(Ours)} &   L  &  -   &     \textbf{69.60} & \textbf{86.75} & \textbf{92.40} & \textbf{52.74} 
\\
\hline
\end{tabular}}
\end{center}
\end{table}

\begin{table*}[!htbp]
\begin{center}

\caption{Ablation studies. "G" and "L" stands for global-matching and local-matching inference protocols.}
\label{ablation}
\resizebox{1.0\textwidth}{!}{
\begin{tabular}{c|c|cccc|cccc}
\hline
\multirow{2}{*}{No.} & \multirow{2}{*}{Methods}  & \multicolumn{4}{c|}{CUHK-PEDES}        & \multicolumn{4}{c}{ICFG-PEDES}        \\
\cline{3-10}
                     &           & Rank-1 & Rank-5 & Rank-10 & mAP & Rank-1 & Rank-5 & Rank-10 & mAP\\
                  \hline\hline
0&Baseline            & 64.36 & 83.36 & 88.78 & 58.18 & 56.16 & 73.77 & 80.17 & 31.59   \\
1&+NDF                     &71.79 & 88.78 & 93.28 & 64.77 & 60.68 & 78.55 & 84.61 & 36.48 \\
2&+NDF+ARA                  & 72.93 & 89.20 & 93.30 & 65.27 & 61.15 & 78.82 & 84.59 & 36.67  \\
3&+NDF+ATP(G)                  & 73.01 & 88.94 & 93.46 & 65.49 & 62.16 & 78.89 & 84.65 & 36.60 \\
4&+NDF+ATP(L)         & 77.78 & 91.15 & 94.50 & 68.46 & 67.09 & 81.91 & 86.85 & 38.99 \\
5&+NDF+ARA+ATP(G)           & 73.48 & 89.57 & 94.10 & 65.82 & 62.54 & 79.46 & 85.14 & 37.07 \\
\hline
6&+NDF+ARA+ATP(L)      & \textbf{78.37} & \textbf{91.57} & \textbf{94.58} & \textbf{68.87} & \textbf{67.81} & \textbf{82.34} & \textbf{87.14} & \textbf{39.85}\\
\hline
\end{tabular}}
\end{center}
\end{table*}

\noindent{\bfseries CUHK-PEDES:} 
The performance comparisons with the related methods on the CUHK-PEDES dataset are shown in Table~\ref{cuhksota}. It can be observed that our proposed CADA achieves state-of-the-art results in both inference protocols. Under the global matching {inference protocol}, CADA reaches 73.48\% Rank-1 accuracy and 65.82\% mAP, which is fairly close to the strongest competitor(IRRA\cite{jiang2023cross}) with a little improvement, i.e.,$+0.10\%$ in Rank-1 accuracy. Under the local-matching {inference protocol}, CADA reaches 78.37\%(+4.99\%) Rank-1 accuracy and 68.87\%(+2.74\%) mAP respectively, outperforming IRRA\cite{jiang2023cross} with a significant margin.
{The results demonstrate that our proposed CADA can better learn the detailed cross-modal associations based on the bidirectional formulation.}
Besides, the global matching results show that building bidirectional local associations contributes in narrowing the modality gap of global representations.

\noindent{\bfseries ICFG-PEDES:} As is shown in Table~\ref{icfgsota}, our proposed CADA obtains results comparable to the state-of-the-art method IRRA\cite{jiang2023cross} under the global-matching {inference protocol}.
Under the local-matching inference protocol, our method outperforms other methods by a large margin, which achieves 67.81\% on Rank-1 accuracy, surpassing the most recent methods IRRA\cite{jiang2023cross} and CFine\cite{yan2022clip} by +4.35\% and +6.98\%. 
Notably, global matching in ICFG-PEDES is insufficient because of the negative impact of complex backgrounds and unstable illumination.
The results demonstrate that CADA could {bidirectionally} learn cross-modal association at the local level to overcome the variance of backgrounds and illumination.

\noindent{\bfseries RSTPReid:} We also conduct experiments on a newly released benchmark dataset RSTPReid and show the results in Table \ref{rstpsota}. It can be observed that our proposed CADA largely outperforms other existing methods, which surpasses IRRA\cite{jiang2023cross} by +9.40\% and +5.57\% on Rank-1, and mAP respectively. Moreover, compared to other local-matching methods, CADA also outperforms all current methods, surpassing the most recent local-matching method CFine\cite{yan2022clip} by +19.05\% on Rank-1 accuracy.

\subsection{Ablation Study}
To fully demonstrate the effectiveness of each module in our CADA framework, we conduct ablation studies on two benchmarks CUHK-PEDES and ICFG-PEDES. {We adopt the dual encoder (i.e., a ViT-B/16 and a BERT-based ) 
with the CMPM~\cite{liu2019deep} loss as the Baseline.}
The experimental results are shown in Table~\ref{ablation}.

\noindent{\bfseries The effectiveness of the Noramlized Distribution Fitting loss.}
Comparing the results of No.0 $vs.$ No.1, we can observe that replacing the commonly used cross-modal projection matching (CMPM) loss with the proposed  $L_{NDF}$, the Rank-1 and mAP are improved by 7.43\% and 6.59\% on CUHK-PEDES, and by 4.52\%, 4.89\% on ICFG-PEDES. The results demonstrate that $L_{NDF}$ is more effective in aligning cross-modal global representations {since it eliminates the effect of vector magnitude variation.} 

\noindent{\bfseries The effectiveness of Association of Text Tokens to Image Patches (ATP).}
The ATP module significantly improves the performance by exploiting the full interaction of text tokens and image patches and building associations. Comparing results in No.1 $vs.$ No.4, it is evident that building token-level associations largely enhances the performance on both precision and recall, obtaining improvement by +4.85\%, +5.94\% on Rank-1 accuracy and by +3.69\%, +2.51\% on mAP. Besides, as ATP learns the cross-modal relation in fine-grained local features, the distance of global features in shared latent space is also narrowed. As revealed in No.1 $vs.$ No.3, even though the decoder is unused in inference, ATP improves Rank-1 accuracy by +1.22\%, +1.48\%. The results demonstrate that during building token-level association, the decoder of CADA learns to infer the matching probability by checking the consistency of local details, meanwhile, the encoders learn to map the global features into latent space more precisely.

\noindent{\bfseries The effectiveness of Association of Image Regions to Text Attributes (ARA).}
{ARA is achieved by the Masked Attribute Modeling, which is used to learn associations from image regions to text attributes.}
In No.1 $vs.$ No.2, it can be observed that directly adding the ARA module into global-matching framework without any token-level interaction, the model gains improvement of +1.14\% and +0.47\% on Rank-1 accuracy, proving that by building up the region-to-attribute association, the performance of unimodal encoders is enhanced and modality gap is narrowed.
By comparing No.3 $vs.$ No.5 and No.4 $vs.$ No.6, we can observe that both global-matching inference and local-matching inference gain improvement from the ARA module. 

\begin{table}[]

\begin{center}
\caption{Evaluation under different combinations of splitting stride and group size in Sec.~\ref{TTA} on CUHK-PEDES.}
\label{group}
\resizebox{0.8\linewidth}{!}{
\begin{tabular}{ccc|cc}
\hline
        
size $p$ & stride $r$ & groups $\kappa$ & Rank-1         & mAP            \\
\hline\hline
36   & 36     & 2      & \textbf{77.78} & \textbf{68.46} \\
36   & 18     & 3      & 77.53          & 68.31          \\
24   & 24     & 3      & 77.45          & 68.45          \\
48   & 24     & 2      & 77.47          & 68.43          \\
72   & 72     & 1      & 77.01          & 68.06 \\
\hline
\end{tabular}}
\end{center}
\end{table}

\begin{figure}[htbp]
	\centering
	\subfloat[Performance under different $\eta$ on the CUHK-PEDES dataset.]{\includegraphics[width=0.8\linewidth]{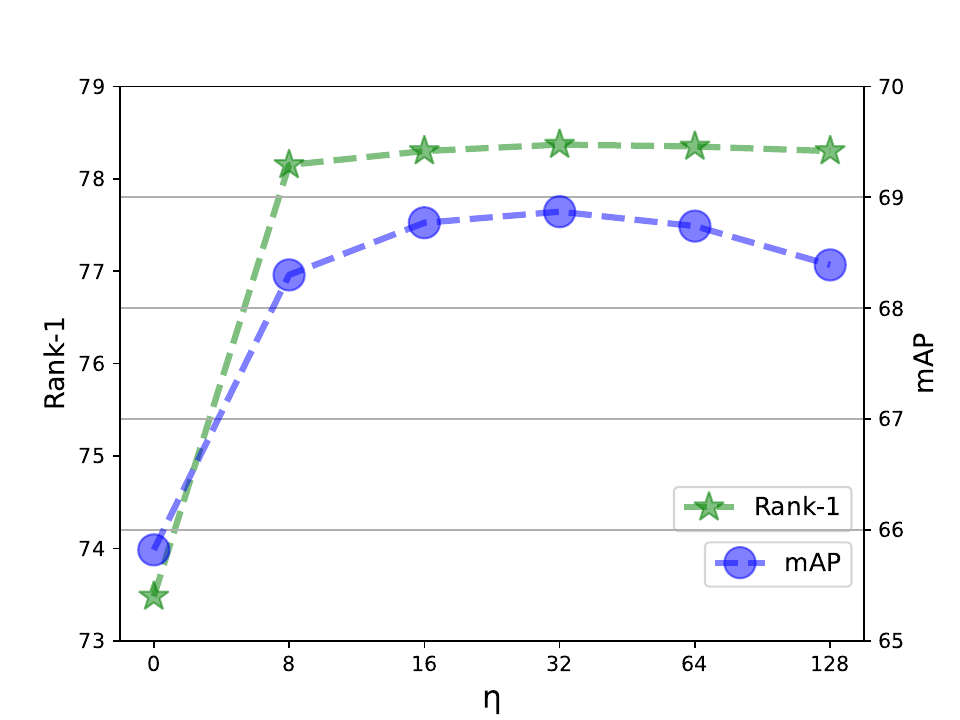}}\hspace{5pt}
	\subfloat[Performance under different $\eta$ on the ICFG-PEDES dataset.]{\includegraphics[width=0.8\linewidth]{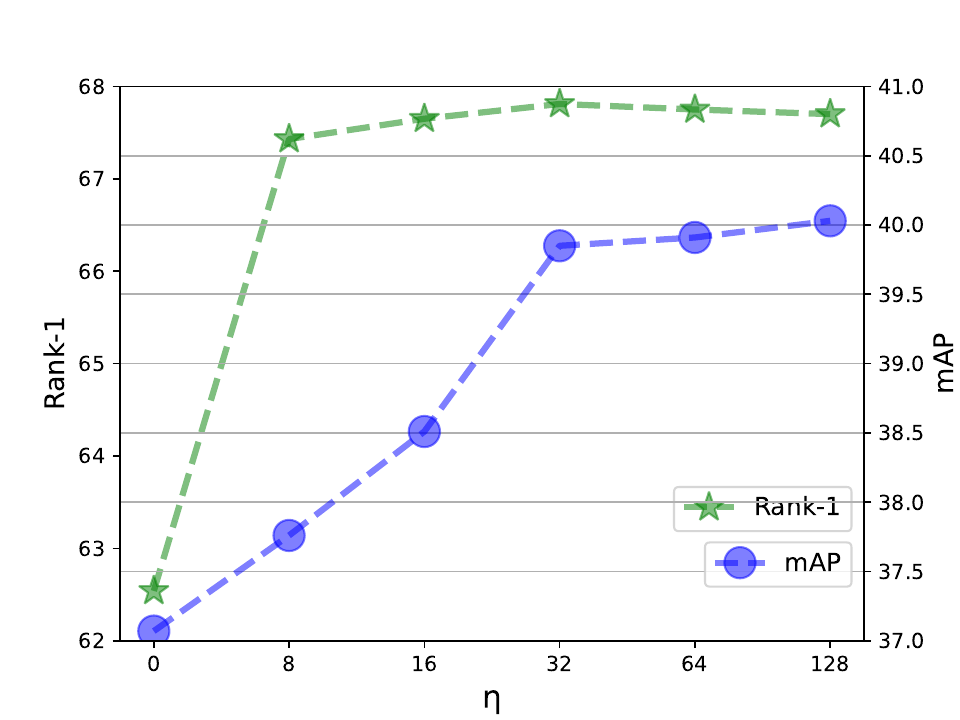}}
  \caption{Performance under different $\eta$.}
  \label{fig:analysis_eta}
\end{figure}

\begin{figure}[!tb]
	\centering
	\centering  
 \setlength{\belowcaptionskip}{-0.5cm}
	\includegraphics[width=0.8\linewidth]{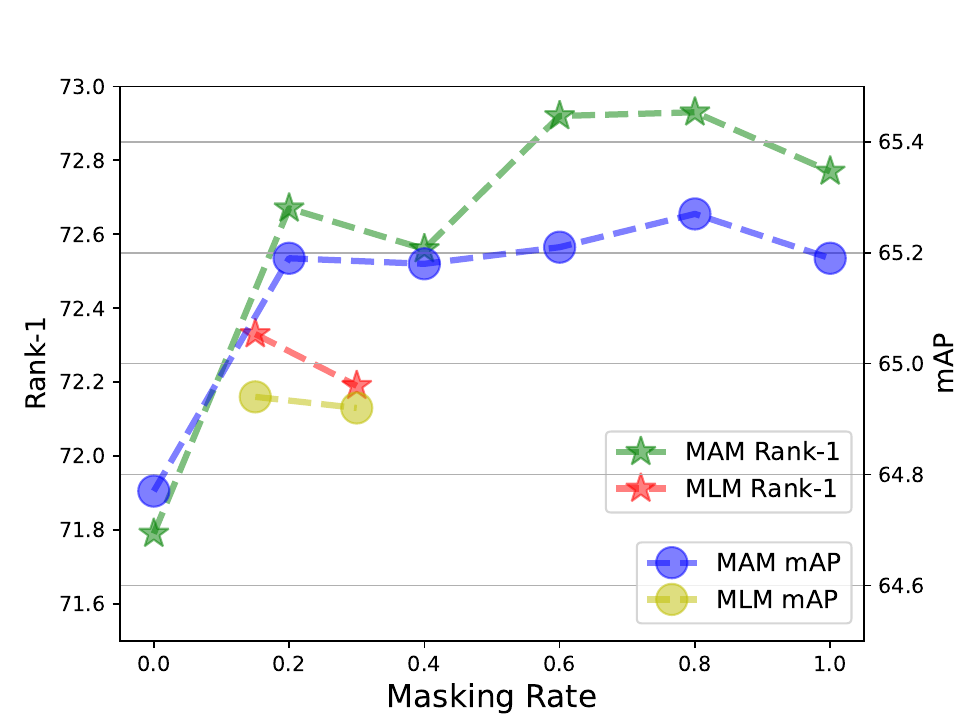}
	  
 \caption{Evaluation on the CUHK-PEDES under different mask rates of MAM. Results of MLM with 0.15 and 0.3 masking rates are also presented as a comparison.}
 \label{fig:analysis_mask}
\end{figure}

\subsection{Hyperparameters Analysis}

\noindent{\bfseries Analysis of local feature grouping.}
{Considering the training efficiency and noises from meaningless
words (e.g., "a", "of", and so on),}
we split local aggregated features into groups to mine local associations in Sec.\ref{TTA} . 
We evaluate how group size $p$ and splitting stride $r$ affect the token-to-patch association. We evaluate our model under different combinations of $p$ and $r$  and report the results in Table~\ref{group}. It can be observed that the performance reaches the best when setting the group size as 32 without overlap. In such splitting method we guarantee each group contains enough local aggregated features to judge the match.

\noindent{\bfseries Analysis of the masking rate of Masked Attribute Modeling. }
We analyze the impact of the MAM masking rate, presented in Sec.\ref{IAA}.
The experimental results of the masking rate setting are shown in Fig.\ref{fig:analysis_mask}. 
We can observe that setting the masking rate to 0.8 can achieve better performance both on Rank1 and mAP metrics. 

\begin{table}[!tbp]
\begin{center}
\caption{Performance comparisons on the domain generalization task. $A \Rightarrow B$ means we take the $A$ dataset as the training set and $B$ dataset as the testing set.}
\label{tab:domaingener}
\resizebox{\linewidth}{!}{
\begin{tabular}{c|ccc|ccc}
\hline
\multirow{2}{*}{Methods} & \multicolumn{3}{c}{CUHK$\Rightarrow$ \text{ICFG}} & \multicolumn{3}{c}{ICFG$\Rightarrow$ \text{CUHK}} \\
                         & Rank-1       & Rank-5       & Rank-10       & Rank-1       & Rank-5       & Rank-10       \\
                         \hline\hline
Dual Path \cite{zheng2020dual}               & 15.41        & 29.80        & 38.19         & 7.63         & 17.14        & 23.52         \\
MIA  \cite{niu2020improving}                    & 19.35        & 36.78        & 46.42         & 10.93        & 23.77        & 32.39         \\
SCAN  \cite{lee2018stacked}                   & 21.27        & 39.26        & 48.43         & 13.63        & 28.61        & 37.05         \\
SSAN   \cite{ding2021semantically}                  & 29.24        & 49.00        & 58.53         & 21.07        & 38.94        & 48.54         \\
LGUR  \cite{shao2022learning}                   & 34.25        & 52.58        & 60.85         & 25.44        & 44.48        & 54.39         \\
\hline
Ours                     & \textbf{52.60} & \textbf{69.03} & \textbf{75.22} & \textbf{54.18} & \textbf{73.68} & \textbf{80.48}     \\
\hline
\end{tabular}}
\end{center}
\end{table}

\noindent {\bfseries Analysis of interaction pairs selection.}
\label{k_test}
In the local-matching inference protocol, we mix the global matching score $S_G$ in Eq.(\ref{eq:globsim}) and the local matching score $S_L$ in Eq.(\ref{eq:localsim}) to improve the matching accuracy. Obviously, simply calculating interaction scores for all image-text pairs brings huge computational costs with a complexity of $O(MN)$, for $M$ images and $N$ texts. Consequently, for each text we only select $\eta$ images which own top $\eta$ global similarity, then conduct $\eta$ times local interaction to obtain their matching scores, shrinking the computational complexity to $O(M+\eta N)$. It is proved by our experiments that $\eta$ can be set pretty low to prevent high computation costs and largely improve performance compared to the global-matching method.
We evaluate our method on the CUHK-PEDES and ICFG-PEDES under different $\eta$, and the experimental results are shown in Fig.\ref{fig:analysis_eta}. We also set $\eta=0$ as a comparison, which means no interaction is conducted. It can be observed that when we set $\eta=32$, we can already get great accuracy improvement, which largely saves computation costs. It is worth noting that even when we limit $\eta<10$, CADA still outperforms all existing state-of-the-art methods, demonstrating the efficiency of our model.

\begin{figure*}[!tb]
    \centering
    \setlength{\belowcaptionskip}{-0.3cm}
    \includegraphics[width=0.9\textwidth]{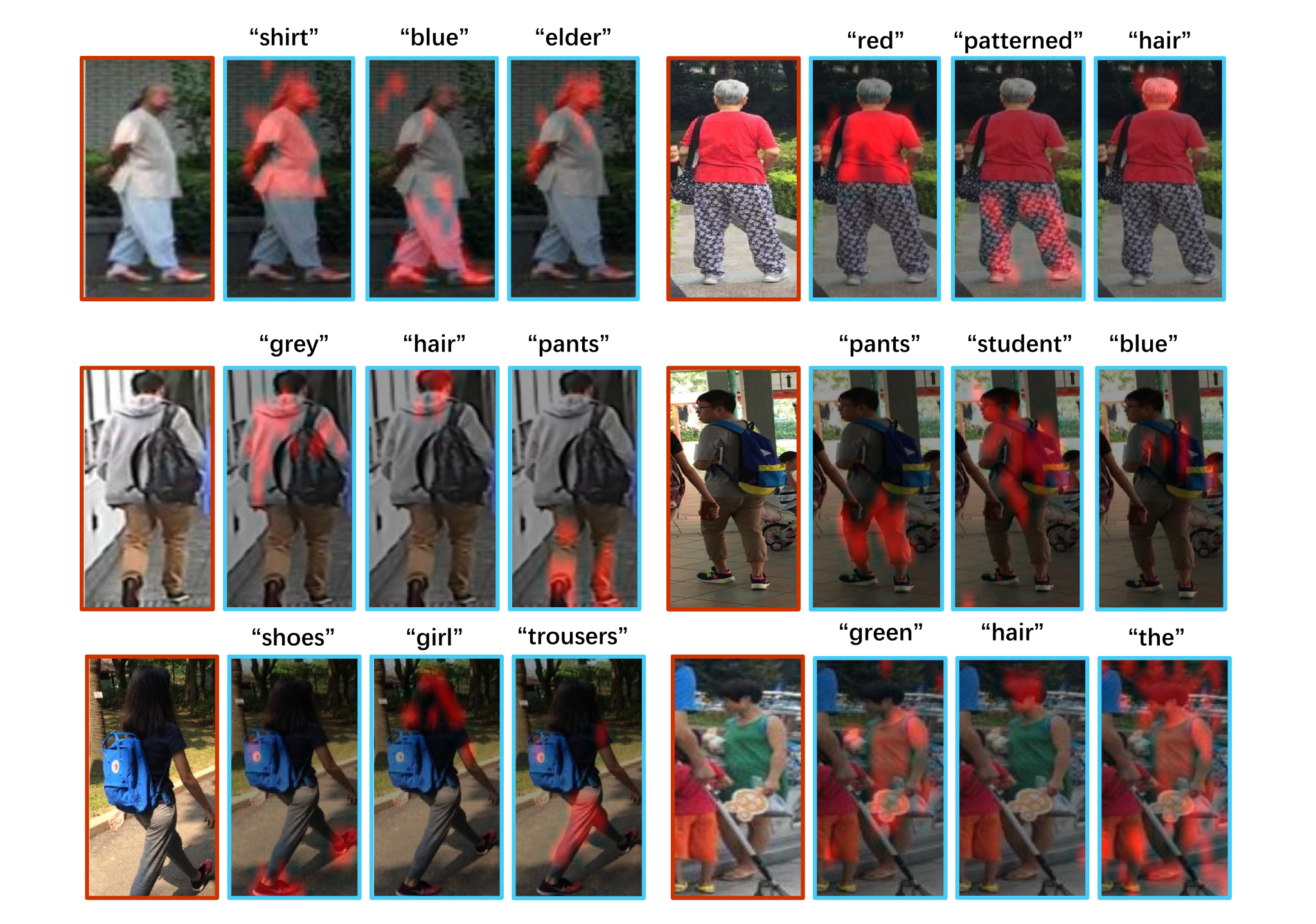}
    \caption{Examples of text tokens to image patches attentions. For each person, we show the image in the red frame and token-to-patch attention images in the blue frame. The attention weight is marked with red dots. }
    \label{fig:token2patch}
\end{figure*}

\begin{figure*}[t]
	\centering
 \setlength{\abovecaptionskip}{-0.3cm}
 \includegraphics[width=0.9\textwidth]{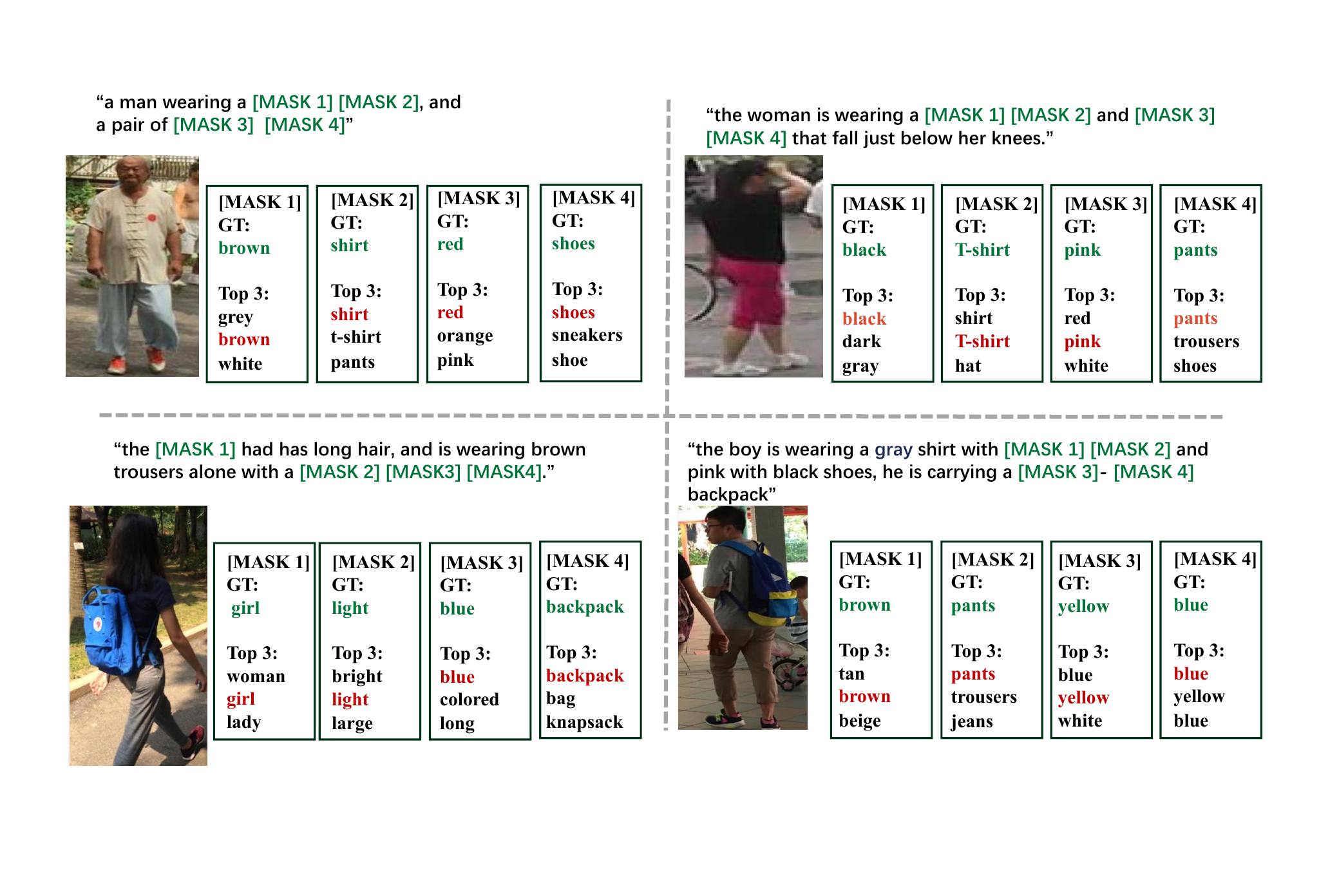}
	\caption{Examples of predicted attributes, we show ground-truth tokens as "GT" and words of Top3 probability. }
	\label{fig:mask_pred}
\end{figure*}

\subsection{Analysis of the Number of Parameters}
In our method, we share the parameters in the text encoder and cross-modal decoder and employ a total of 36 attention layers in our model: 12 self-attention layers in the Vision encoder and 12 self-attention layers + 12 cross-attention layers in the textual encoder.

In order to evaluate the impact of increasing parameters, we conducted further experiments using different numbers of attention layers and report the number of parameters as shown in Table~\ref{tab:domaingener}. The results demonstrate that reducing the number of attention layers leads to a corresponding drop in performance.
However, it is worth noting that when we reduce the layers to 24, our model contains fewer parameters than IRRA\cite{jiang2023cross} while still outperforming it.

In summary, our experimental results demonstrate that 
%
(1) increasing the number of parameters indeed improves performance.
(2) Our improvement is not simply because of the increasing parameters since our method still outperforms IRRA \cite{jiang2023cross} even when holding the same number of parameters.

\begin{table}[!tbp]
\begin{center}
\caption{Comparison of the performance and quantity of parameters. We show the results of our model under a different number of attention layers.}
\label{tab:domaingener}
\resizebox{0.9\linewidth}{!}{
\begin{tabular}{l|ll|l}
\hline
\multicolumn{1}{c|}{\multirow{2}{*}{methods}} & \multicolumn{2}{c|}{Rank-1} & \multicolumn{1}{c}{\multirow{2}{*}{param}}   \\
\cline{2-3}
\multicolumn{1}{c|}{}                         & ICFG         & CUHK        & \multicolumn{1}{c}{}                         \\
\hline \hline
Ours (36layers)   & 67.81   & 78.37   & 223.45M \\
Ours (32layers) &   67.13      & 77.18   & 204.55M \\
Ours (28layers) & 66.63   & 76.17   & 185.65M \\
Ours (24layers) & 64.35   & 75.09   & 166.74M \\
IRRA  \cite{jiang2023cross}          & 63.46   & 73.38   & 190.43M \\
TIPCB   \cite{chen2022tipcb}         & 54.96   & 64.26   & 184.75M \\
\hline
\end{tabular}}
\end{center}
\end{table}

\subsection{Qualitative Results}

Fig.~\ref{fig:token2patch} illustrates the attention weights of image patches for a given text word. It clearly shows that for color-related words (e.g. "red", and "blue"), our CADA precisely captures the image patches in the corresponding color. For words related to clothing or body parts (e.g. "pants", "hair"), CADA adaptively pays attention to the whole relevant region. We also show the attention map of the image-irrelevant word (e.g. "the"), and get an irregular attention map, revealing that these words can be confusing for building image-text associations, which demonstrates the effectiveness of the association of text Tokens to image patches(ATP).

In Fig.\ref{fig:mask_pred} we illustrate the influence of the association of image regions to text attributes (ARA) module in our proposed approach.
For each masked attribute word, we show the top-3 predictions {in Eq.(\ref{maskPred})}.
Given cues from image patches, our model successfully builds correspondence from visual information to attributes e.g. gender, color, clothing, and body parts in descriptive text.

\subsection{More Evaluation}
\noindent {\bfseries Comparisons on the domain generalization task.} Our proposed CADA effectively learns association in fine-grained details. Due to its effectiveness in building associations between fine-grained details, the impact from variations of background or illusion can be subtle. To further verify the generalization ability of CADA, we conduct a domain generalization analysis on two benchmarks, where we directly employ CADA pre-trained on the source benchmark to evaluate on target benchmark. The results are reported in Table \ref{tab:domaingener}. Our proposed CADA outperforms all methods by a large margin. For example, our CADA outperforms LGUR\cite{shao2022learning} by +18.35\% and +28.47\% on Rank-1 accuracy.

\section{Conclusion}
In this work, {we point out that the associations between text and images are relevant to the anchors and develop a cross-modal adaptive dual association (CADA) framework.}
Our framework consists of a global-level association and a decoder-based adaptive dual association module. The module enables bidirectional and adaptive cross-modal correspondence associations between visual and textual modalities through the Association of text Tokens to image Patches (ATP) and the Association of image Regions to text Attributes (ARA). Specifically, in ATP, we propose passing information from text tokens to image patches to build token-level associations. In contrast, in ARA, we propose generating masked text phrases by adaptively locating related image regions. By building a bidirectional cross-modal association, we achieve remarkable performance on all three popular benchmarks, demonstrating the effectiveness of our approach.

\bibliographystyle{IEEEtran}
\bibliography{cada}

\end{document}